\documentclass[letterpaper]{article} %
\usepackage[]{aaai24}  %
\usepackage{times}  %
\usepackage{helvet}  %
\usepackage{courier}  %
\usepackage[hyphens]{url}  %
\usepackage{graphicx} %
\usepackage{natbib}  %
\usepackage{caption} %
\usepackage{algorithm}
\usepackage{algorithmic}

\usepackage{newfloat}
\usepackage{listings}
\DeclareCaptionStyle{ruled}{labelfont=normalfont,labelsep=colon,strut=off} %
\floatstyle{ruled}
\newfloat{listing}{tb}{lst}{}
\floatname{listing}{Listing}
\title{Exploiting Language Models as a Source of Knowledge for Cognitive Agents}
\author{
    James R. Kirk, Robert E. Wray, John E. Laird
}
\affiliations{
    Center for Integrated Cognition @ IQMRI\\

    24 Frank Lloyd Wright Drive
Suite H 1200, Box 464 \\
Ann Arbor, Michigan 48106\\
    \{james.kirk,robert.wray,john.laird@cic.iqmri.org\}
}

\nocopyright

\begin{document}

\maketitle

\begin{abstract}
Large language models (LLMs) provide capabilities far beyond sentence completion, including question answering, summarization, and natural-language inference. While many of these capabilities have potential application to cognitive systems, our research is exploiting language models as a source of task knowledge for cognitive agents, that is, agents realized via a cognitive architecture. We identify challenges and opportunities for using language models as an external knowledge source for cognitive systems and possible ways to improve the effectiveness of knowledge extraction by integrating extraction with cognitive architecture capabilities, highlighting with examples from our recent work in this area.  
\end{abstract}

\section{Introduction}

Cognitive architectures\footnote{Portions of this paper draw from and update  \citet{wray_language_2021}.} \cite{anderson_integrated_2004,laird_soar_2012,kotseruba_40_2020,newell_unified_1990} are foundational tools  for research toward the realization of cognitive capabilities. Architectures also facilitate development of cognitive systems (``agents'') that manifest various integrated cognitive capabilities, including planning, learning, plan execution, and many others. An agent uses these capabilities in concert to act, to perform sophisticated tasks, and to achieve short- and long-term goals.

A key limitation of scaling agents to ever larger and more complex tasks and applications is their ability to acquire and integrate new task knowledge. As a consequence, a significant thrust of applied cognitive architecture research over the years has been exploring various knowledge engineering  \cite{yost_acquiring_1993,crossman_high_2004,ritter_high-level_2006}, experiential learning \cite{nejati_learning_2006,choi_evolution_2018,pearson_toward_1998}, and instructional \cite{huffman_flexibly_1995,gluck_interactive_2019} approaches intended to mitigate the costs of acquiring new, effective, and robust task knowledge. While improvements have been achieved, none of these approaches has resulted in routine, low-cost, large-scale knowledge resources for cognitive agents. 

In contrast, Large Language Models (LLMs) \cite{openai_gpt-4_2023,driess_palm-e_2023} provide a huge breadth of potential knowledge. However, exploiting this knowledge is challenging because the production of knowledge is unreliable and untrustworthy \citep{lenat_getting_2023}. As a knowledge source, LLM responses can be corrupted by hallucination, irrelevancy, incorrectness and can be unethical and/or unsafe. Further, for most applications of LLMs, the models themselves are fixed: they do not learn or adapt to the specific situations in which they are used.

We are exploring the hypothesis that an integration of LLMs and cognitive architectures can offset the limitations of each in the context of reliable knowledge scaling. LLMs can be used as a rich, (largely) low-cost source of knowledge for agents to extend their task knowledge. In concert, agent cognitive capabilities, realized within a cognitive architecture, can be applied to the LLM to improve the ``precision" and correctness of the task knowledge that an LLM is asked to produce. 
This approach emphasizes what cognitive architectures do best (support end-to-end integration of interaction, reasoning, language processing, learning, etc., using structured, curated knowledge) and what language models do best (provide associational retrieval from massive stores of latent unstructured, possibly unreliable knowledge). We expect that by using LLMs to acquire new task knowledge, agents can reduce their reliance on more costly sources of task knowledge (e.g., extensive training, human instruction, explicit knowledge engineering) and, consequently, scale more readily to larger task domains and applications.

In this paper, we discuss the dimensions of this problem, requirements for a solution, and some examples of some interactions between LLMs and cognitive architectures that appear important for task acquisition. 

\section{Potential Patterns of Integration}

\begin{figure}[t]
    \centering
    \includegraphics[width=.9\columnwidth]{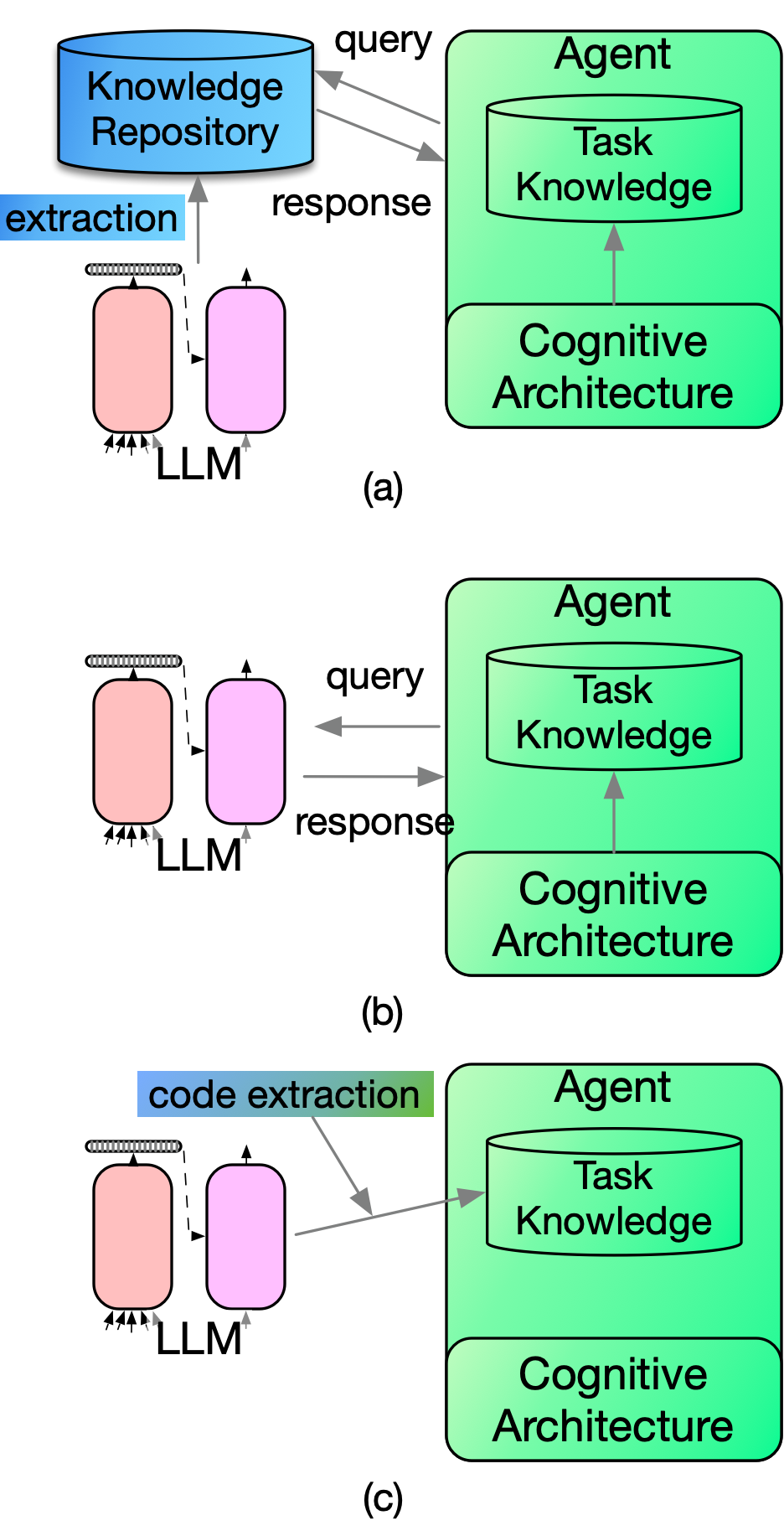}
    \caption{Possible approaches to CA-LLM integration in support of task-knowledge acquisition.}
    \label{fig:integration-patterns}
\end{figure}

We introduce three different ways cognitive-architecture-LLM integration could be realized for acquiring task knowledge, and we outline potential benefits and costs of each.\footnote{This list is not comprehensive and other options for considering LLMs as source of knowledge for cognitive agents are feasible; e.g., see  \citet{lenat_getting_2023} for an alternative list.} The alternatives are illustrated in Figure~\ref{fig:integration-patterns}. In the diagrams, an agent is viewed as a combination of its (cognitive) architecture and content (task knowledge), the standard formulation of an agent \cite{russell_artificial_1995}. While this structure (over-)simplifies the details of most cognitive agents, it is a reasonable approximation for considering systems-integration alternatives.

In the figure, we assume LLMs (and in 1a, external knowledge repositories) as sources of general or unspecialized knowledge. In some cases, for a particular task, domain- or task-specific knowledge bases may be available or an LLM that has been fine-tuned for a specific problem domain. When these are available, an agent should generally attempt to use them. However, we are more focused on the case of a general intelligent agent where the agent (or developer) cannot assume that a special-purpose knowledge source is readily available. In this case, the agent generally will need to attempt to extract knowledge from more general resources, like an LLM or a world-knowledge repository such as ConceptNet \cite{speer_conceptnet_2017}.

The three integration patterns we consider are:

\begin{itemize}

    \item (a) \textbf{Indirect extraction:} In this option,  general extraction processes designed for LLMs are used and responses from an LLM are placed in a knowledge store. The agent then accesses the knowledge store to obtain task knowledge. For example, general knowledge extraction from LLMs is being developed to populate knowledge graphs \cite{bosselut_comet_2019} which is a specific example of the use of LLMs as a knowledge base \cite{petroni_language_2019}. As a consequence, this alternative leverages that ongoing research. Further, any existing cognitive agent capabilities that draw and exploit external knowledge stores could also be re-purposed for exploring this alternative \cite{wray_language_2021}.
    \item (b) \textbf{Direct extraction:} In this option, the agent directly formulates and sends queries to the LLM and then interprets the responses it receives. The responses are interpreted and internalized within with agent's processing, resulting in situation-specific learning of new task knowledge (represented by the arrow from the cognitive architecture to task knowledge components). This option requires that the agent encode capabilities that perform the same kinds of processes needed for indirect extraction (i.e., the blue ``extraction" process in 1a, but within its own agent knowledge).
    \item (c) \textbf{Direct Knowledge Encoding:} This option provides ``direct wiring" for the agent,  encoding task knowledge directly into the agent's internal knowledge representations and memories. For instance, using program-code generation capabilities of LLMs \cite{chen_evaluating_2021,austin_program_2021}, researchers have shown that LLMs can be used to create programs for robotic control of embodied agents \cite{singh_progprompt_2023,brohan_rt-2_2023}. Because cognitive architectures generally provide programmable interfaces, code generation might be apt for cognitive-architecture agents. This option could thus employ an external extraction process that programmed agent knowledge directly. However, while these extraction processes might be able to build on existing extraction methods, the extraction processes would need to be specialized for cognitive architectures (and likely for individual architectures).
  
\end{itemize}

Our work to-date focuses on direct extraction. We chose direct extraction over indirect extraction for two reasons. First, in indirect extraction (1a), task-knowledge extraction is not tied to the agent's specific, current knowledge needs. We hypothesize that direct extraction enables the agent to use its specific situation and context (including its embodiment) to query the LLM resulting in improved precision (relevance to the specific situation). Second, because extraction processes not tied to the agent cannot fully anticipate agent needs, the knowledge base in 1a cannot be assumed to be an exhaustive resource for all the knowledge that could be extracted from the LLM. Thus, even an agent that used 1a might need to sometimes resort to processes akin to 1b when the knowledge base was found lacking.

The primary potential downside of the direct extraction approach is that the agent must have the ability to manage the potential challenges of extraction from the LLM itself (e.g., determining whether some response is plausible) rather than depending on specialized extraction processes (as in 1a). In other words, the agent must have more sophisticated capabilities to directly query and interpret LLM responses.

Direct encoding is another option that we recommend be explored in the community, although we do not discuss it further here. Because task specification languages and tools have already been developed (for both general AI systems and cognitive agents), it may be feasible to leverage the code generation capabilities of LLMs to generate executable task knowledge. Fine-tuning a code generation model with examples that illustrate how task capabilities are realized within the distinct representational and memory systems of cognitive architectures is likely to result in better results than in-context learning because the execution models of cognitive architectures often differ substantially from traditional programming languages.

\section{Direct Extraction: \\ Challenges and Opportunities}

Knowledge extraction is the process by which an agent gains knowledge of its task and/or environment from an external knowledge source. Successful extraction results in the agent having (new) knowledge it can bring to bear on its tasks. From a cognitive-systems perspective, what is important is that the knowledge produced by an extraction process is ``actionable” by the agent. Thus, the goal of extraction is not simply to add knowledge but to add knowledge that allows the agent to improve its ability to function as an autonomous entity in a multitask environment. 

Direct extraction requires that an agent directly interact with an LLM (send queries, receive responses) rather than through an intermediary. In this section, we characterize some of the features of LLMs relevant to direct extraction and, in some cases, contrast those extraction processes with extraction from more curated knowledge sources (e.g., Cyc, WordNet, ConceptNet). Relevant features include:

\begin{itemize}
    \item Breadth and Depth of Knowledge: A major strength of LLMs, such as GPT3, GPT4 or PALM-E, compared to curated knowledge bases (KBs) is their extensive breadth of encoded knowledge. 
    \item Provenance and Accuracy of Knowledge: The quality of behavior for a cognitive agent is invariably tied to the quality of knowledge it reasons with. With traditional KBs, task knowledge is typically either curated or at least derived from the agent’s own experience with the world. In contrast, LLMs are (largely) derived from uncurated web resources, and the knowledge's provenance is unknown and very likely includes errors and conflicts. 
    \item Relevance of Knowledge: Even when the LLM contains the knowledge relevant to the agent’s needs, extracting it can be highly sensitive to the specific query sent to the LLM \cite{pezeshkpour_large_2023}. In addition to the basic question, a query to an LLM will often include additional context \cite{reynolds_prompt_2021}, related examples \cite{brown_language_2020}, and additional instructions, such as ``show your reasoning steps" \cite{wang_self-consistency_2023}. The resulting responses are highly dependent on all these factors. This sensitivity to the query context makes it difficult to ensure that whatever information is retrieved from the LLM is actually relevant to the context of the agent. 
    \item Situatedness of Knowledge: Curated knowledge bases (such as Cyc) and LLMs encode ``ungrounded" knowledge about the world. As general-purposes resources, they do not encode knowledge about an agent’s current situation, its embodiment (what it can sense and how it can act), and its goals and plans, which may be encoded in an agent’s long-term semantic or episodic memory. Thus, an open question is the extent to which the agent can embed relevant information from its understanding of its situation to obtain knowledge that can be connected or ``grounded" to its performance context. As mentioned above, the potential disconnect between the general knowledge of an LLM and the specific knowledge needs of an agent motivates the exploration of direct extraction methods in particular.
    \item Accessibility of Knowledge: In a typical AI knowledge base, the APIs for query/response and knowledge representation are well-defined, making it straightforward for an agent to attempt to retrieve information and to parse any responses. For an LLM, the specific form of a request and response are (generally) less structured, e.g., natural-language sentences. While tools such as LangChain's output parser\footnote{https://github.com/langchain-ai/langchain} are designed to bridge accessibility issues, it remains a challenge for an agent to interpret responses from the LLM; in the extreme, an agent must parse natural language to extract what information is provided in a response. 
    \item Structural Integration: Traditional knowledge bases impose low to moderate computational costs and latency and have reliable access. In contrast, many LLMs, especially the largest, are web resources with restricted access, high relative latency, and  access depends on internet connectivity. Many LLMs charge by the token sent and received, imposing direct economic constraints on efficient and robust interaction.    
    A cognitive agent will thus need to be strategic in using an LLM in applications that involve real-time environmental and human interaction.
    
\end{itemize}

\section{Requirements and Measures}

For our research, we have pursued a strategy that prioritizes the issues of Accuracy, Relevance, Situatedness, and Accessibility from the previous section. For direct extraction to be practical, an embodied agent needs methods that allow it to elicit responses from the LLM in a manner that produces highly relevant ones (when an LLM has the capacity to potentially produce almost anything), that are appropriate given the situation, and that the agent can access or interpret (which is required for any internalization and use in actual task execution). Here, we further define and refine four requirements \cite{kirk_improving_2023} that an agent must meet to successfully extract actionable task knowledge from an LLM. Specifically, responses from the LLM must be:

\begin{enumerate}

\item \textbf{Interpretable:} To use responses, the agent must be able to parse and understand them. Responses that are represented as code can often be directly executed by an agent; e.g., JSON, first-order logic \cite{olmo_gpt3--plan_2021}, action commands as in RT2 \cite{brohan_rt-2_2023} or even Python programs \cite{singh_progprompt_2023}. More typically, natural language (NL) responses from an LLM require NLP capabilities that the agent can use to interpret those responses. Because responses from the LLM must be interpretable given the agent's existing capabilities (as above), the agent's query construction process must attempt to guide LLM responses so that the resulting responses conform to whatever those native agent language capabilities are.

\item \textbf{Groundable to Situation:} As above, an agent must determine if/how it can situate an LLM response to its current environment. Objects, properties, relations, actions, etc. referenced in the response must be connected to the agent's situation. In the simplest case, groundedness requires the agent to map from perceptions to responses.

\item \textbf{Compatible with Affordances and Embodiment:} The LLM response must also be compatible with the agent's embodiment and the affordances known to the agent. For example, an agent might be embodied in a robot with a single arm or as a phone app acting as a personal assistant. When the affordances or embodiment of the agent are not comparable to those of humans, eliciting compatible responses may be particularly challenging for an LLM trained on a corpus (typically and primarily) describing human activities that presume human embodiment and affordances.

\item \textbf{Match Human Expectations:} Users will have differing preferences about how tasks or actions should be performed and what behavior is appropriate for different situations \cite{kirk_improving_2022}. This criterion is important because it is impossible for the LLM (alone) to provide responses that ensure a match to human preferences. For example, imagine a household robotic agent tasked with putting away groceries. Should a can of beans be stored in a cupboard or a pantry? Either answer is plausible \cite{lenat_getting_2023}. However, in some specific home, there is likely a clear preference for one location for the beans to be stored. The resulting requirement, from the point of view of extraction, is that an agent must anticipate the need to elicit human preferences because the LLM cannot disambiguate between plausible alternatives when the user has a preference for one or another.

\end{enumerate}

The first three requirements are related to the \textit{viability} of extracted task knowledge; in other words, is the agent able to interpret, internalize, and use a response in service of a task? The fourth requirement derives from constraints that arise from particular task environments: responses should be aligned with human expectations. 

As above, the threshold for the fourth requirement is that the responses produced by the LLM must be plausible. A response that was viable, but that suggested that the can beans from the previous example should be stored in the sink or dish rack is very likely to be wrong. For our research, the more important criterion is that the response should be not just reasonable or plausible but also responsive and relevant to a specific human user's expectations for that situation.

Agents must evaluate LLM responses to identify if responses meet these requirements. When these requirements are not met, if the agent attempts to use them (as is), the result will be incorrect learning or failure. 

These requirements also suggest how we can, as researchers, measure and evaluate progress toward direct extraction. What fraction of responses were viable when attempting to extract knowledge in support of a new task? For those that were not viable, what fraction failed due to incompatibility with affordances vs. interpretability, etc? Measuring performance on these kinds of questions can facilitate the evaluation of progress and comparison of alternative approaches.

\begin{figure}[t]
    \centering
    \includegraphics[width=.9\columnwidth]{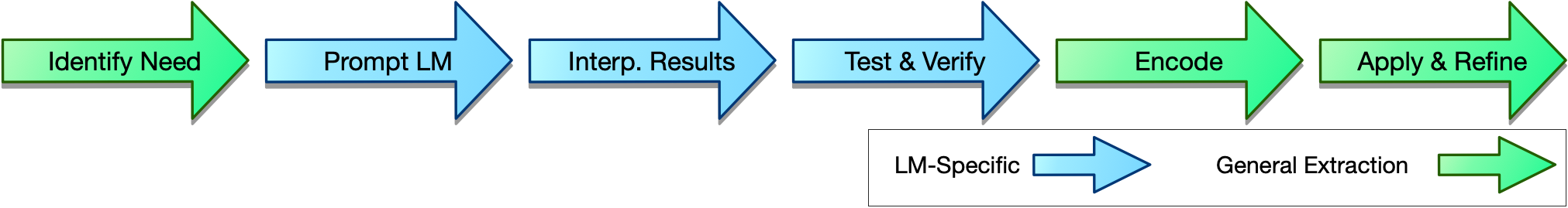}
    \caption{Schematic process for direct extraction of task knowledge for task learning in a cognitive agent.}
    \label{fig:processing-steps}
\end{figure}

\section{Supporting Direct Extraction}
In this section, we briefly summarize our high-level approach to direct extraction. We have used this approach to enable agents to learn new tasks, demonstrating the feasibility of the direct extraction approach \cite{kirk_improving_2022,kirk_improving_2023,kirk_integrating_2023}. 

Figure~\ref{fig:processing-steps} illustrates the basic process. Rather than extracting knowledge for only the sake of gaining new knowledge, the cognitive agent seeks to extract knowledge for a specific purpose: enabling the agent to perform a (current or anticipated) task. Our extraction strategy takes advantage of prior work in cognitive systems in accessing and using external knowledge sources (some of these steps are also used for internal knowledge access). 

Steps that overlap with general extraction patterns are illustrated in green; process steps specific to LLMs are in blue. Our current implementations use an existing agent \cite{kirk_learning_2019,mininger_expanding_2021} that learns from human instruction. These components provide general extraction capabilities that can be employed for LLMs (i.e., the agent is already ``extracting" knowledge from a human instructor via natural-language interaction). In what follows, we refer to this agent and its capabilities as the ``Interactive Task Learning Agent" (or ITL Agent). We have extended this ITL Agent, which previously obtained new knowledge from natural-language interaction with a user, to now also extract knowledge thru interaction with an LLM. Roughly, the extraction process comprises the following six steps:

\begin{enumerate}
    \item The agent identifies a knowledge need, such as a gap in its knowledge. The ITL agent already has existing capabilities to detect knowledge gaps (including assessing what kinds of gaps it has encountered, such as needing a goal or action, not understanding a term, etc.) In our current work, we assume that the agent seeks the LLM first as a resource for potentially providing the ability to bridge these knowledge gaps and then relies on human input only when the LLM is insufficient. (Longer term, cognitive agents will likely need to evaluate if/when an LLM is an appropriate source for a specific gap.) 
    
    Importantly, from the point of view of the integration patterns described in Section 2, the agent's fine-grained identification of a gap or need makes direct extraction highly relevant to resolving the gap.  The agent can use its specific need to develop a more precise and targeted query to the LLM. We employ a template-based prompting approach, a common method in prompt engineering \cite{reynolds_prompt_2021}. Templates are focused on specific gaps that the agent may encounter \cite{wray_language_2021}, such as eliciting the steps needed to perform a task \cite{kirk_improving_2022}, eliciting task subgoals, such as putting away the can of beans as part of tidying the kitchen \cite{kirk_integrating_2023}, and refining a prior response \cite{kirk_improving_2023}.

\item The agent prompts an LLM, choosing a specific prompt template based on its evaluation of its knowledge gap/need. It instantiates the template with information/data it has from the situation and context, and sets parameters to the LLM given the situation (e.g., setting the temperature or ``variability" in the LLM response). Thus, the resulting, agent-constructed prompt is not only specific to the context but also the particular type of knowledge gap. (We further detail this approach to prompting in a subsequent section.)

\item The agent then interprets the response(s) from the LLM. Although we did briefly explore having an LLM generate logical expressions \cite{kirk_improving_2022}, interpretation for our approach requires the agent to use its internal natural-language understanding capabilities to convert the LLM response text to the agent’s internal knowledge representation. Because the ITL Agent already has natural-language processing (NLP) capabilities for learning from human instruction, the extended approach leverages that existing NLP capability. However, this choice also limits what the agent can interpret because the ITL Agent's NLP capabilities are fairly modest. To date, one of the primary pain points in our explorations has been determining if/when an interpretation problem should be resolved by further refining LLM responses to match our agent's NLP capabilities or whether to change/refine the existing NLP capabilities.

\item Because the results from the LLM are not necessarily accurate and reliable (as noted above), the agent evaluates, tests, and attempts to verify extracted results from the LLM. Although verification of acquired knowledge has been explored in cognitive systems, verification of knowledge derived from LLMs presents new challenges. For example, in the ITL Agent, we assumed that the human instructor would provide accurate, grounded, interpretable knowledge to the agent that reflected the instructor's preferences. None of these assumptions are likely to consistently hold for a response from an LLM, resulting in verification processes that must address all the requirements introduced above. We have developed a novel, multi-step approach to verification, detailed further below.

\item Following verification, the agent encodes the knowledge it has obtained into its own memory(-ies) as appropriate for current and future use in task performance. In the ITL Agent, this encoding is accomplished via a 2-step process. First, the agent takes the step(s) indicated by a verified LLM response (such as planning to achieve a goal or executing an action), allowing the agent to determine in practice (in addition to the analytic assessment during verification) if the new knowledge suits the task. A second step involves the agent deliberately reviewing the steps of the task (``retrospection"). Using Soar's chunking process \cite{laird_chunking_1986}, the retrospective analysis is compiled into new procedural knowledge that enables the agent to perform the same step/task in the future without resorting to the LLM. While other approaches to encoding could be realized (even others within Soar), this approach, first developed for the ITL Agent, is sufficient for encoding new knowledge that it derives from the LLM. 

\item In the final step, the agent uses the knowledge it has acquired and continues to monitor its correctness and utility, and refines it based on its experience in using it. Although it is likely that knowledge could be incomplete, over-general, etc., to date, we have not encountered significant problems that require downstream refinement and thus have not yet explored if LLMs present unique new requirements for this step in the process.

\end{enumerate}

\section{Prompting Approach}
The agent must choose a prompting strategy and then formulate a specific prompt appropriate for its task and environment, the knowledge needed, and the requirements of the LLM itself. Prompt engineering \cite{reynolds_prompt_2021}, crafting and refining prompts so that they produce  desired results, has been shown to be an effective strategy for retrieving reasonable knowledge. Using template-based prompting, one kind of prompt engineering, an agent selects an appropriate template designed to elicit the desired knowledge and then instantiates the template with context specific to the task it is attempting to learn.\footnote{This approach is used so routinely that it is now directly supported in LLM development tools such as LangChain.} To date, we have used a template-based approach, although we have only had to develop a few templates, as shown in Table~\ref{tab:prompt-templates}.

\begin{table}[tb]
\begin{tabular}{lp{.73\columnwidth}}
\hline
\textbf{Template}        & \textbf{Description Example}                                     \\ \hline \hline
Goal & Request a goal for a given task, situation, 
and object of focus. \\
Action & Request the next action step given a task, situation, object of focus, and steps thus far.\\
Repair & Request a re-formulation of a response, given the previous prompt, the failed response, and a categorization of the type of failure. \\

\hline
\end{tabular}
\caption{Examples of templates used in template-based prompting.}
\label{tab:prompt-templates}
\end{table}

We combine template-based prompting with few-shot prompting 
\cite{brown_language_2020}. Few-shot prompting embeds examples of desired responses in the prompt. Prompt examples include similar queries and the respective desired response to influence the LLM's response to the main prompt. One of the main roles examples play in our approach is biasing responses toward simple and direct language that the ITL Agent's NLP interpreter can parse. Prompt examples can also introduce patterns for the LLM to follow, such as in Chain of Thought \cite{wei_chain--thought_2022}, which influence the LLM to reason about problems step by step. 

\begin{table}[t]
    \centering
    \begin{tabular}{l}
    \hline
    (EXAMPLES)\\
    ~~~(TASK)Task name: store object. \\ ~~~~~Task context: I am in mailroom. 
Aware of package  \\ ~~~~~ of office supplies; package is in mailroom.\\
~~~ (RESULT) The goal is that the package is in the closet \\
~~~~~ and the closet is closed.(END RESULT) \\
~~~(END TASK) \\
~~~(TASK)Task name: deliver package. \\ ~~~~~ Task context: I am in mailroom. 
Aware of package \\ ~~~~~ addressed to Gary; package is in mailroom. \\
~~~~~ (RESULT)The goal is that the package is in Gary’s office. \\~~~~~ (END RESULT) 
\\ ~~~(END TASK) \\
(END EXAMPLES) \\
(TASK)Task name: \textbf{tidy kitchen}. \\ ~~~ Task context: I am in \textbf{kitchen}. \\
~~~ Aware of \textbf{mug} in \textbf{dish rack.}\\ ~~~(RESULT) \\
\hline

    \end{tabular}
    \caption{Example of an agent-created prompt for eliciting goals. Agent instantiations in the prompt from its situational context are highlighted in \textbf{bold}.}
    \label{tab:prompt_example}
\end{table}

This discussion provides a high-level outline of our template-based prompting approach which is detailed elsewhere \cite{kirk_improving_2022}. Table~\ref{tab:prompt_example} presents an example of a prompt constructed by the agent for a task to ``tidy a kitchen" in which the agent is looking at a mug in a dish rack in that kitchen.\footnote{The indentation is for human readability alone; the prompt is constructed without line breaks.} In our work to-date, we have primarily used GPT-3 \cite{brown_language_2020} as the LLM for research. From the point of view of direct extraction, the relatively simple approach enables the agent to construct prompts that effectively elicit mostly interpretable (and viable) responses.

\section{Verification Approach}

Another burgeoning research area is identifying effective tools to verify the responses of LLMs. These include ranking responses from the LLM based on interaction with and feedback from the environment \citet{logeswaran_few-shot_2022},  response sampling \cite{wang_self-consistency_2023}, using planning knowledge \cite{valmeekam_planning_2023}, additional LLM prompting about the veracity of retrieved responses \cite{kim_language_2023}, and using human feedback/annotation \cite{wu_tidybot_2023,kirk_improving_2022}.

\begin{figure}[t]
    \centering
    \includegraphics[width=1.0\columnwidth]{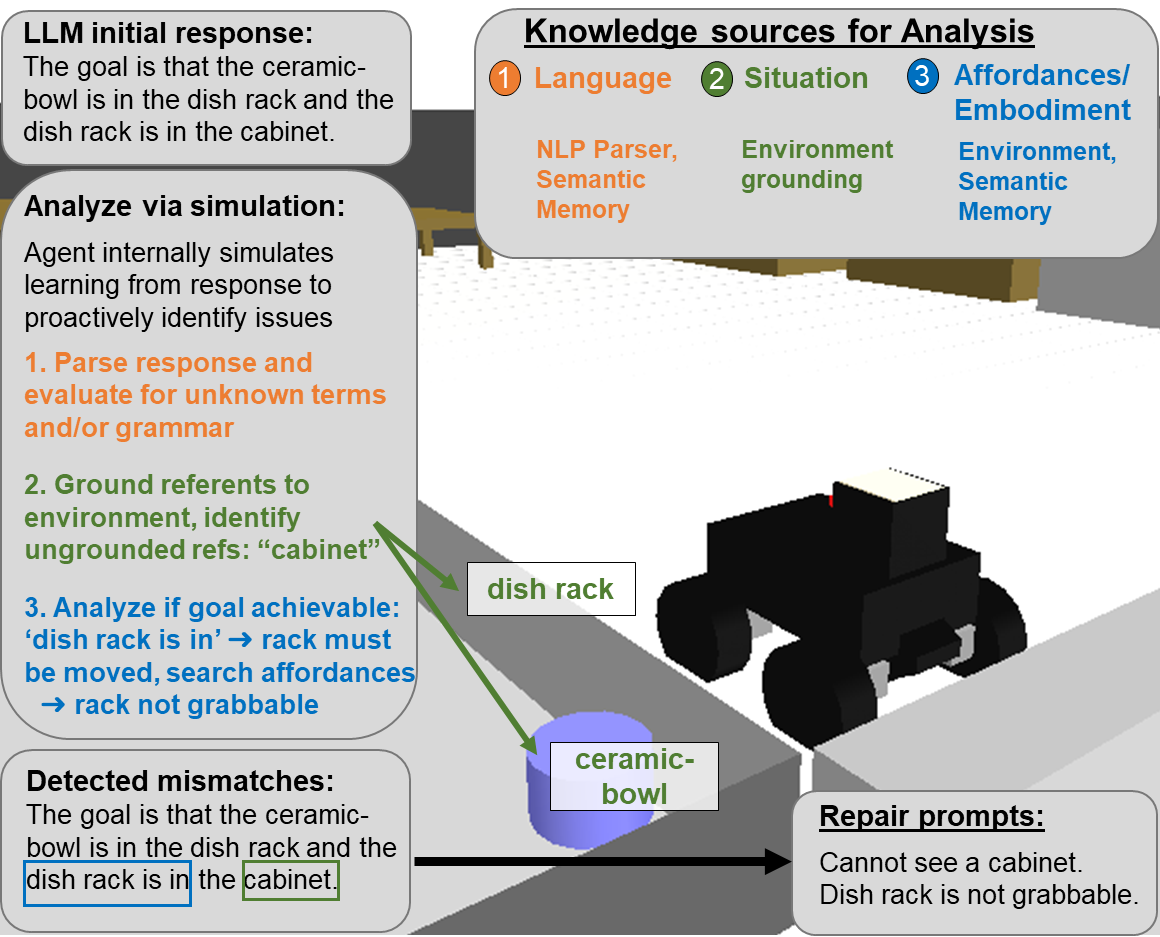}
    \caption{Agent analysis of LLM responses via internal simulation}
    \label{fig:analysis}
\end{figure}

Cognitive architectures provide both a framework for the evaluation of LLM responses and the knowledge (encoded in various memories) required to analyze them. As outlined briefly above, our agent simulates the process of learning from a response in order to evaluate the result of using that response \cite{kirk_improving_2023}. 

Figure \ref{fig:analysis} summarizes the primary components of this analysis and the relevant memories from the cognitive architecture that the agent relies on for analysis. The agent uses its NLP parser and linguistic knowledge encoded in semantic memory to evaluate if the response is interpretable by the agent (1, orange). It uses knowledge of the current situation (encoded in working memory) to ground references in the LLM response and to evaluate if any references cannot be grounded (2, green). Finally, it uses knowledge encoded in semantic memory, and the context of the current environment from working memory, to analyze if responses align with its embodiment and affordances (3, blue), evaluating if the task goal is achievable by the agent. 

Once the analysis is complete, any issues that are identified can be used in subsequent prompts to repair the responses that are misaligned with these requirements. The repair template was (outlined in Table~\ref{tab:prompt-templates}) is comparable to the prompt shown in Table~\ref{tab:prompt_example} but adds the incorrect LLM response, the identified issue (e.g., a word that is unknown), and asks for another response.

A final strategy for evaluation is enabling human oversight by asking a user if a task action or goal is correct before the agent uses the response. Human evaluation enables the agent to conform to the final requirement, aligning with human expectations. Correct task performance often requires eliciting individual human preferences, as discussed above.

\begin{figure}[t]
    \centering
    \includegraphics[width=1.0\columnwidth]{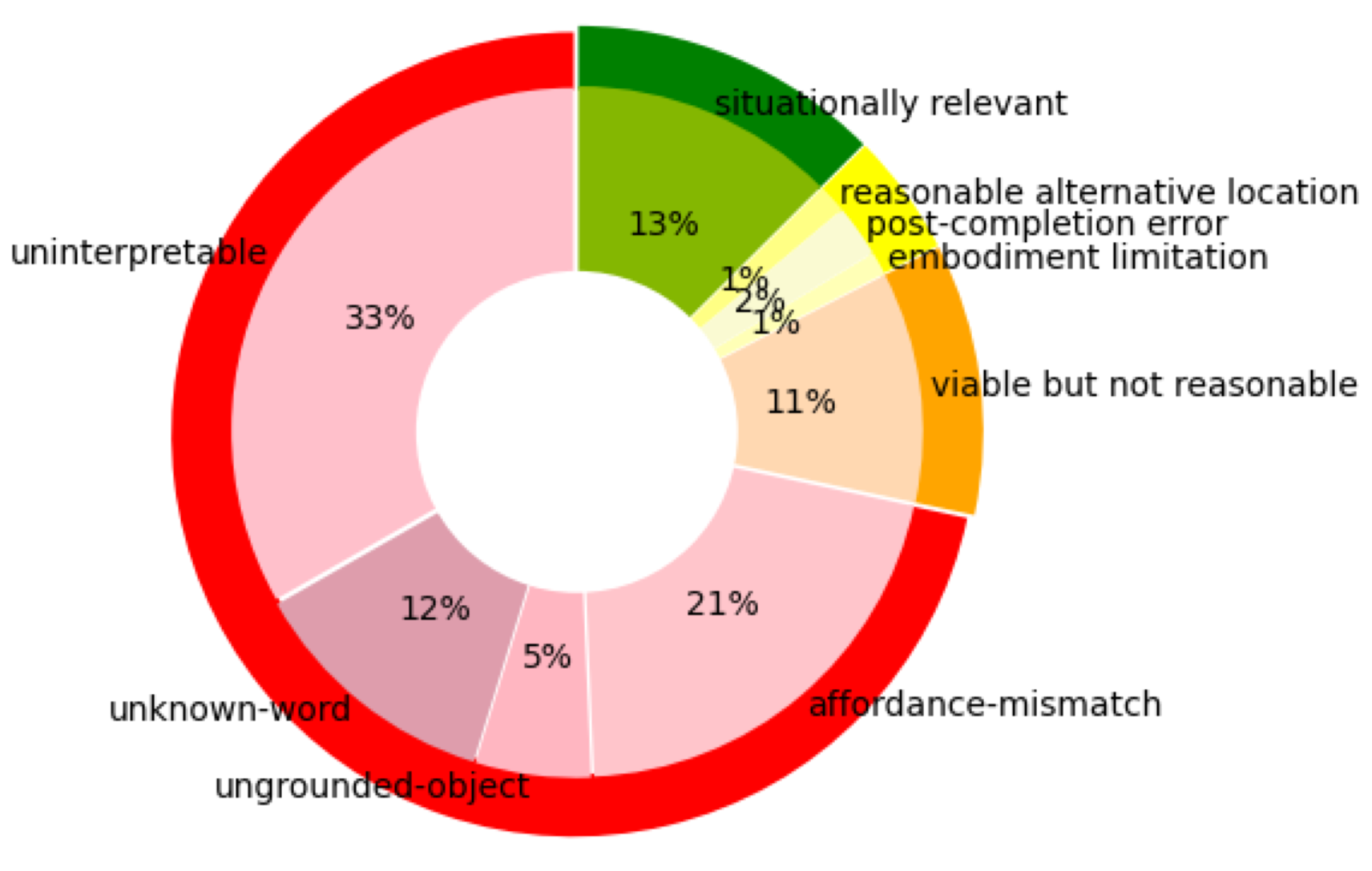}
    \caption{Categorization of responses retrieved from the LLM during agent experiment.}
    \label{fig:response_categories}
\end{figure}

Figure \ref{fig:response_categories} shows an analysis of responses extracted from the LLM by our agent during an experiment where it learns to tidy a kitchen with 35 common kitchen objects. The chart shows the (human-determined) classification of all the responses retrieved from the LLM, including unviable responses in red (not aligned with the first three requirements), viable but not reasonable responses in orange, reasonable responses in yellow, and situationally relevant responses in green that match the human preferences for this task. A takeaway from this analysis is the large percentage of total responses (over 70\%) that are not viable for the embodied agent, indicating the necessity for evaluation of responses for reliable learning. In other words, this iterative prompt-refine-re-prompt approach to verification allows the agent to generate and to identify the relatively small proportion of responses that are viable and situationally relevant, resulting in ``actionable" knowledge for the agent. 

Further evaluation of responses using other capabilities of cognitive architectures is potentially useful, but not yet explored. A cognitive architecture agent could use episodic memory to see if retrieved knowledge matches actions performed in the past. It could also use planning knowledge to see if retrieved goals are achievable, or retrieved actions are executable. Cognitive architectures also support interfacing with other knowledge sources (e.g., knowledge bases such as WordNet or ConceptNet) which could provide additional information for evaluation (e.g., finding synonyms for unknown words).

\section{Conclusion}

Autonomous systems, whether they are realized with cognitive architectures or not, will have to acquire new knowledge to perform tasks and accomplish their goals. 
However, the lack of reliable, scalable acquisition of new task knowledge, especially online acquisition of knowledge, has limited the operation and impact of cognitive systems. The integration of LLMs with cognitive architectures presents an intriguing opportunity to exploit the breadth of knowledge in LLMs to overcome limits on knowledge acquisition.

In this paper, we presented various ways one might approach this problem and highlighted the potential of direct extraction from LLMs as an integration path. We summarized the challenges and requirements for exploring this integration and a high-level, step-wise process for pursuing this goal. We outlined some of the ways we are attempting to pursue this research vision, highlighting the use of template-based prompting and knowledge-driven evaluation that enables more reliable and useful responses from the LLM. 

A more complete realization of the entire task-learning pipeline (as envisioned in Figure~\ref{fig:processing-steps}), as well as an evaluation of the pipeline in terms of scaling for knowledge acquisition, remain as future work. One notable result in terms of scaling, however, has been to observe the synergistic interactions between different sources of knowledge within task learning. The extended ITL Agent uses look-ahead planning, human oversight, and the LLM to attempt to acquire new knowledge. Early results \cite{kirk_integrating_2023} suggest that planning can virtually eliminate the need for an agent to ask for actions (at least in the task domains we have explored) when the agent acquires a correct (i.e., verified) goal description. Similarly, using LLMs to elicit goals in conjunction with the verification process requires significantly less human oversight. In summary, this integrated-knowledge approach realized within and enabled by a cognitive architecture, is suggestive of a potential breakthrough in knowledge acquisition and task learning for cognitive agents.

\bibliography{calm-refs,custom}

\begin{thebibliography}{40}
\providecommand{\natexlab}[1]{#1}

\bibitem[{Anderson et~al.(2004)Anderson, Bothell, Byrne, Douglass, Lebiere, and
  Qin}]{anderson_integrated_2004}
Anderson, J.~R.; Bothell, D.; Byrne, M.~D.; Douglass, S.; Lebiere, C.; and Qin,
  Y. 2004.
\newblock An integrated theory of the mind.
\newblock \emph{Psychological review}, 111(4): 1036.

\bibitem[{Austin et~al.(2021)Austin, Odena, Nye, Bosma, Michalewski, Dohan,
  Jiang, Cai, Terry, Le, and Sutton}]{austin_program_2021}
Austin, J.; Odena, A.; Nye, M.; Bosma, M.; Michalewski, H.; Dohan, D.; Jiang,
  E.; Cai, C.; Terry, M.; Le, Q.; and Sutton, C. 2021.
\newblock Program {Synthesis} with {Large} {Language} {Models}.
\newblock ArXiv:2108.07732 [cs].

\bibitem[{Bosselut et~al.(2019)Bosselut, Rashkin, Sap, Malaviya, Celikyilmaz,
  and Choi}]{bosselut_comet_2019}
Bosselut, A.; Rashkin, H.; Sap, M.; Malaviya, C.; Celikyilmaz, A.; and Choi, Y.
  2019.
\newblock {COMET}: {Commonsense} {Transformers} for {Automatic} {Knowledge}
  {Graph} {Construction}.
\newblock In \emph{Proceedings of the 57th {Annual} {Meeting} of the
  {Association} for {Computational} {Linguistics}}.
\newblock ArXiv: 1906.05317.

\bibitem[{Brohan et~al.(2023)Brohan, Brown, Carbajal, Chebotar, Chen,
  Choromanski, Ding, Driess, Dubey, Finn, Florence, Fu, Arenas, Gopalakrishnan,
  Han, Hausman, Herzog, Hsu, Ichter, Irpan, Joshi, Julian, Kalashnikov, Kuang,
  Leal, Lee, Lee, Levine, Lu, Michalewski, Mordatch, Pertsch, Rao, Reymann,
  Ryoo, Salazar, Sanketi, Sermanet, Singh, Singh, Soricut, Tran, Vanhoucke,
  Vuong, Wahid, Welker, Wohlhart, Wu, Xia, Xiao, Xu, Xu, Yu, and
  Zitkovich}]{brohan_rt-2_2023}
Brohan, A.; Brown, N.; Carbajal, J.; Chebotar, Y.; Chen, X.; Choromanski, K.;
  Ding, T.; Driess, D.; Dubey, A.; Finn, C.; Florence, P.; Fu, C.; Arenas,
  M.~G.; Gopalakrishnan, K.; Han, K.; Hausman, K.; Herzog, A.; Hsu, J.; Ichter,
  B.; Irpan, A.; Joshi, N.; Julian, R.; Kalashnikov, D.; Kuang, Y.; Leal, I.;
  Lee, L.; Lee, T.-W.~E.; Levine, S.; Lu, Y.; Michalewski, H.; Mordatch, I.;
  Pertsch, K.; Rao, K.; Reymann, K.; Ryoo, M.; Salazar, G.; Sanketi, P.;
  Sermanet, P.; Singh, J.; Singh, A.; Soricut, R.; Tran, H.; Vanhoucke, V.;
  Vuong, Q.; Wahid, A.; Welker, S.; Wohlhart, P.; Wu, J.; Xia, F.; Xiao, T.;
  Xu, P.; Xu, S.; Yu, T.; and Zitkovich, B. 2023.
\newblock {RT}-2: {Vision}-{Language}-{Action} {Models} {Transfer} {Web}
  {Knowledge} to {Robotic} {Control}.
\newblock ArXiv:2307.15818 [cs].

\bibitem[{Brown et~al.(2020)Brown, Mann, Ryder, Subbiah, Kaplan, Dhariwal,
  Neelakantan, Shyam, Sastry, Askell, Agarwal, Herbert-Voss, Krueger, Henighan,
  Child, Ramesh, Ziegler, Wu, Winter, Hesse, Chen, Sigler, Litwin, Gray, Chess,
  Clark, Berner, McCandlish, Radford, Sutskever, and
  Amodei}]{brown_language_2020}
Brown, T.; Mann, B.; Ryder, N.; Subbiah, M.; Kaplan, J.~D.; Dhariwal, P.;
  Neelakantan, A.; Shyam, P.; Sastry, G.; Askell, A.; Agarwal, S.;
  Herbert-Voss, A.; Krueger, G.; Henighan, T.; Child, R.; Ramesh, A.; Ziegler,
  D.; Wu, J.; Winter, C.; Hesse, C.; Chen, M.; Sigler, E.; Litwin, M.; Gray,
  S.; Chess, B.; Clark, J.; Berner, C.; McCandlish, S.; Radford, A.; Sutskever,
  I.; and Amodei, D. 2020.
\newblock Language {Models} are {Few}-{Shot} {Learners}.
\newblock In Larochelle, H.; Ranzato, M.; Hadsell, R.; Balcan, M.~F.; and Lin,
  H., eds., \emph{Advances in {Neural} {Information} {Processing} {Systems}},
  volume~33, 1877--1901. Curran Associates, Inc.

\bibitem[{Chen et~al.(2021)Chen, Tworek, Jun, Yuan, Pinto, Kaplan, Edwards,
  Burda, Joseph, Brockman, Ray, Puri, Krueger, Petrov, Khlaaf, Sastry, Mishkin,
  Chan, Gray, Ryder, Pavlov, Power, Kaiser, Bavarian, Winter, Tillet, Such,
  Cummings, Plappert, Chantzis, Barnes, Herbert-Voss, Guss, Nichol, Paino,
  Tezak, Tang, Babuschkin, Balaji, Jain, Saunders, Hesse, Carr, Leike, Achiam,
  Misra, Morikawa, Radford, Knight, Brundage, Murati, Mayer, Welinder, McGrew,
  Amodei, McCandlish, Sutskever, and Zaremba}]{chen_evaluating_2021}
Chen, M.; Tworek, J.; Jun, H.; Yuan, Q.; Pinto, H. P. d.~O.; Kaplan, J.;
  Edwards, H.; Burda, Y.; Joseph, N.; Brockman, G.; Ray, A.; Puri, R.; Krueger,
  G.; Petrov, M.; Khlaaf, H.; Sastry, G.; Mishkin, P.; Chan, B.; Gray, S.;
  Ryder, N.; Pavlov, M.; Power, A.; Kaiser, L.; Bavarian, M.; Winter, C.;
  Tillet, P.; Such, F.~P.; Cummings, D.; Plappert, M.; Chantzis, F.; Barnes,
  E.; Herbert-Voss, A.; Guss, W.~H.; Nichol, A.; Paino, A.; Tezak, N.; Tang,
  J.; Babuschkin, I.; Balaji, S.; Jain, S.; Saunders, W.; Hesse, C.; Carr,
  A.~N.; Leike, J.; Achiam, J.; Misra, V.; Morikawa, E.; Radford, A.; Knight,
  M.; Brundage, M.; Murati, M.; Mayer, K.; Welinder, P.; McGrew, B.; Amodei,
  D.; McCandlish, S.; Sutskever, I.; and Zaremba, W. 2021.
\newblock Evaluating {Large} {Language} {Models} {Trained} on {Code}.
\newblock ArXiv:2107.03374 [cs].

\bibitem[{Choi and Langley(2018)}]{choi_evolution_2018}
Choi, D.; and Langley, P. 2018.
\newblock Evolution of the {Icarus} {Cognitive} {Architecture}.
\newblock \emph{Cognitive Systems Research}, 48: 25--38.

\bibitem[{Crossman et~al.(2004)Crossman, Wray, Jones, and
  Lebiere}]{crossman_high_2004}
Crossman, J.; Wray, R.~E.; Jones, R.~M.; and Lebiere, C. 2004.
\newblock A {High} {Level} {Symbolic} {Representation} for {Behavior}
  {Modeling}.
\newblock In Gluck, K., ed., \emph{Proceedings of 2004 {Behavior}
  {Representation} in {Modeling} and {Simulation} {Conference}}. Arlington, VA.

\bibitem[{Driess et~al.(2023)Driess, Xia, Sajjadi, Lynch, Chowdhery, Ichter,
  Wahid, Tompson, Vuong, Yu, Huang, Chebotar, Sermanet, Duckworth, Levine,
  Vanhoucke, Hausman, Toussaint, Greff, Zeng, Mordatch, and
  Florence}]{driess_palm-e_2023}
Driess, D.; Xia, F.; Sajjadi, M. S.~M.; Lynch, C.; Chowdhery, A.; Ichter, B.;
  Wahid, A.; Tompson, J.; Vuong, Q.; Yu, T.; Huang, W.; Chebotar, Y.; Sermanet,
  P.; Duckworth, D.; Levine, S.; Vanhoucke, V.; Hausman, K.; Toussaint, M.;
  Greff, K.; Zeng, A.; Mordatch, I.; and Florence, P. 2023.
\newblock {PaLM}-{E}: {An} {Embodied} {Multimodal} {Language} {Model}.
\newblock ArXiv:2303.03378 [cs].

\bibitem[{Gluck, Laird, and Lupp(2019)}]{gluck_interactive_2019}
Gluck, K.; Laird, J.; and Lupp, J.~R., eds. 2019.
\newblock \emph{Interactive {Task} {Learning}: {Agents}, {Robots}, and {Humans}
  {Acquiring} {New} {Tasks} through {Natural} {Interactions}}, volume~26 of
  \emph{Strüngmann {Forum} {Reports}}.
\newblock Cambridge, MA: MIT Press.

\bibitem[{Huffman and Laird(1995)}]{huffman_flexibly_1995}
Huffman, S.~B.; and Laird, J.~E. 1995.
\newblock Flexibly instructable agents.
\newblock \emph{Journal of Artificial Intelligence Research}, 3: 271--324.

\bibitem[{Kim, Baldi, and McAleer(2023)}]{kim_language_2023}
Kim, G.; Baldi, P.; and McAleer, S. 2023.
\newblock Language models can solve computer tasks.
\newblock ArXiv:2303.17491 [cs].

\bibitem[{Kirk et~al.(2022)Kirk, Wray, Lindes, and Laird}]{kirk_improving_2022}
Kirk, J.; Wray, R.~E.; Lindes, P.; and Laird, J.~E. 2022.
\newblock Improving {Language} {Model} {Prompting} in {Support} of
  {Semi}-autonomous {Task} {Learning}.
\newblock In \emph{Proceedings of the 2022 {Advances} in {Cognitive} {Systems}
  {Conference}}. Washington, DC/Virtual.

\bibitem[{Kirk and Laird(2019)}]{kirk_learning_2019}
Kirk, J.~R.; and Laird, J.~E. 2019.
\newblock Learning {Hierarchical} {Symbolic} {Representations} to {Support}
  {Interactive} {Task} {Learning} and {Knowledge} {Transfer}.
\newblock In \emph{Proceedings of {IJCAI} 2019}, 6095--6102. International
  Joint Conferences on Artificial Intelligence.

\bibitem[{Kirk, Wray, and Lindes(2023)}]{kirk_improving_2023}
Kirk, J.~R.; Wray, R.~E.; and Lindes, P. 2023.
\newblock Improving {Knowledge} {Extraction} from {LLMs} for {Task} {Learning}
  through {Agent} {Analysis}.
\newblock ArXiv:2306.06770 [cs].

\bibitem[{Kirk et~al.(2023)Kirk, Wray, Lindes, and
  Laird}]{kirk_integrating_2023}
Kirk, J.~R.; Wray, R.~E.; Lindes, P.; and Laird, J.~E. 2023.
\newblock Integrating {Diverse} {Knowledge} {Sources} for {Online} {One}-shot
  {Learning} of {Novel} {Tasks}.
\newblock ArXiv:2208.09554 [cs].

\bibitem[{Kotseruba and Tsotsos(2020)}]{kotseruba_40_2020}
Kotseruba, I.; and Tsotsos, J.~K. 2020.
\newblock 40 years of cognitive architectures: core cognitive abilities and
  practical applications.
\newblock \emph{Artificial Intelligence Review}, 53(1): 17--94.

\bibitem[{Laird(2012)}]{laird_soar_2012}
Laird, J.~E. 2012.
\newblock \emph{The {Soar} {Cognitive} {Architecture}}.
\newblock Cambridge, MA: MIT Press.

\bibitem[{Laird, Rosenbloom, and Newell(1986)}]{laird_chunking_1986}
Laird, J.~E.; Rosenbloom, P.~S.; and Newell, A. 1986.
\newblock Chunking in {Soar}: {The} anatomy of a general learning mechanism.
\newblock \emph{Machine Learning}, 1(1): 11--46.

\bibitem[{Lenat and Marcus(2023)}]{lenat_getting_2023}
Lenat, D.; and Marcus, G. 2023.
\newblock Getting from {Generative} {AI} to {Trustworthy} {AI}: {What} {LLMs}
  might learn from {Cyc}.
\newblock ArXiv:2308.04445 [cs].

\bibitem[{Logeswaran et~al.(2022)Logeswaran, Fu, Lee, and
  Lee}]{logeswaran_few-shot_2022}
Logeswaran, L.; Fu, Y.; Lee, M.; and Lee, H. 2022.
\newblock Few-shot {Subgoal} {Planning} with {Language} {Models}.
\newblock In \emph{Proceedings of the {NAACL} 2022}. arXiv.
\newblock ArXiv:2205.14288 [cs].

\bibitem[{Mininger(2021)}]{mininger_expanding_2021}
Mininger, A. 2021.
\newblock \emph{Expanding {Task} {Diversity} in {Explanation}-{Based}
  {Interactive} {Task} {Learning}}.
\newblock Ph.{D}. {Thesis}, University of Michigan, Ann Arbor.

\bibitem[{Nejati, Langley, and Konik(2006)}]{nejati_learning_2006}
Nejati, N.; Langley, P.; and Konik, T. 2006.
\newblock Learning hierarchical task networks by observation.
\newblock In \emph{Proceedings of the 23rd international conference on
  {Machine} learning}, {ICML} '06, 665--672. New York, NY, USA: Association for
  Computing Machinery.
\newblock ISBN 978-1-59593-383-6.

\bibitem[{Newell(1990)}]{newell_unified_1990}
Newell, A. 1990.
\newblock \emph{Unified {Theories} of {Cognition}}.
\newblock Cambridge, Massachusetts: Harvard University Press.

\bibitem[{Olmo, Sreedharan, and Kambhampati(2021)}]{olmo_gpt3--plan_2021}
Olmo, A.; Sreedharan, S.; and Kambhampati, S. 2021.
\newblock {GPT3}-to-plan: {Extracting} plans from text using {GPT}-3.
\newblock In \emph{Proc. of {ICAPS} {FinPlan} and {ICAPS} {KEPS}}.
\newblock ArXiv: 2106.07131.

\bibitem[{OpenAI(2023)}]{openai_gpt-4_2023}
OpenAI. 2023.
\newblock {GPT}-4 {Technical} {Report}.
\newblock ArXiv:2303.08774 [cs].

\bibitem[{Pearson and Laird(1998)}]{pearson_toward_1998}
Pearson, D.~J.; and Laird, J.~E. 1998.
\newblock Toward incremental knowledge correction for agents in complex
  environments.
\newblock \emph{Machine Intelligence}, 15.

\bibitem[{Petroni et~al.(2019)Petroni, Rocktäschel, Lewis, Bakhtin, Wu,
  Miller, and Riedel}]{petroni_language_2019}
Petroni, F.; Rocktäschel, T.; Lewis, P.; Bakhtin, A.; Wu, Y.; Miller, A.~H.;
  and Riedel, S. 2019.
\newblock Language {Models} as {Knowledge} {Bases}?
\newblock In \emph{{EMNLP} 2019}.
\newblock ArXiv: 1909.01066.

\bibitem[{Pezeshkpour and Hruschka(2023)}]{pezeshkpour_large_2023}
Pezeshkpour, P.; and Hruschka, E. 2023.
\newblock Large {Language} {Models} {Sensitivity} to {The} {Order} of {Options}
  in {Multiple}-{Choice} {Questions}.
\newblock ArXiv:2308.11483 [cs].

\bibitem[{Reynolds and McDonell(2021)}]{reynolds_prompt_2021}
Reynolds, L.; and McDonell, K. 2021.
\newblock Prompt {Programming} for {Large} {Language} {Models}: {Beyond} the
  {Few}-{Shot} {Paradigm}.
\newblock In \emph{Extended {Abstracts} of the 2021 {CHI} {Conference} on
  {Human} {Factors} in {Computing} {Systems}}, {CHI} {EA} '21, 1--7. New York,
  NY, USA: Association for Computing Machinery.
\newblock ISBN 978-1-4503-8095-9.

\bibitem[{Ritter et~al.(2006)Ritter, Haynes, Cohen, Howes, John, Best, Lebiere,
  Lewis, St.~Amant, McBraide, Urbas, Leuchter, and
  Vera}]{ritter_high-level_2006}
Ritter, F.~E.; Haynes, S.~R.; Cohen, M.; Howes, A.; John, B.; Best, B.;
  Lebiere, C.; Lewis, R.~L.; St.~Amant, R.; McBraide, S.~P.; Urbas, L.;
  Leuchter, S.; and Vera, A. 2006.
\newblock High-level behavior representation languages revisited.

\bibitem[{Russell and Norvig(1995)}]{russell_artificial_1995}
Russell, S.; and Norvig, P. 1995.
\newblock \emph{Artificial {Intelligence}: {A} {Modern} {Approach}}.
\newblock Upper Saddle River, NJ: Prentice-Hall.

\bibitem[{Singh et~al.(2023)Singh, Blukis, Mousavian, Goyal, Xu, Tremblay, Fox,
  Thomason, and Garg}]{singh_progprompt_2023}
Singh, I.; Blukis, V.; Mousavian, A.; Goyal, A.; Xu, D.; Tremblay, J.; Fox, D.;
  Thomason, J.; and Garg, A. 2023.
\newblock {ProgPrompt}: program generation for situated robot task planning
  using large language models.
\newblock \emph{Autonomous Robots}.

\bibitem[{Speer, Chin, and Havasi(2017)}]{speer_conceptnet_2017}
Speer, R.; Chin, J.; and Havasi, C. 2017.
\newblock {ConceptNet} 5.5: an open multilingual graph of general knowledge.
\newblock In \emph{Proc. of the 31st {AAAI} {Conference} on {Artificial}
  {Intelligence}}, {AAAI}'17, 4444--4451. San Francisco, California, USA: AAAI
  Press.

\bibitem[{Valmeekam et~al.(2023)Valmeekam, Sreedharan, Marquez, Olmo, and
  Kambhampati}]{valmeekam_planning_2023}
Valmeekam, K.; Sreedharan, S.; Marquez, M.; Olmo, A.; and Kambhampati, S. 2023.
\newblock On the {Planning} {Abilities} of {Large} {Language} {Models} ({A}
  {Critical} {Investigation} with a {Proposed} {Benchmark}).
\newblock ArXiv:2302.06706 [cs].

\bibitem[{Wang et~al.(2023)Wang, Wei, Schuurmans, Le, Chi, Narang, Chowdhery,
  and Zhou}]{wang_self-consistency_2023}
Wang, X.; Wei, J.; Schuurmans, D.; Le, Q.; Chi, E.; Narang, S.; Chowdhery, A.;
  and Zhou, D. 2023.
\newblock Self-{Consistency} {Improves} {Chain} of {Thought} {Reasoning} in
  {Language} {Models}.
\newblock In \emph{{ICLR} 2023}. arXiv.
\newblock ArXiv:2203.11171 [cs].

\bibitem[{Wei et~al.(2022)Wei, Wang, Schuurmans, Bosma, Ichter, Xia, Chi, Le,
  and Zhou}]{wei_chain--thought_2022}
Wei, J.; Wang, X.; Schuurmans, D.; Bosma, M.; Ichter, B.; Xia, F.; Chi, E.; Le,
  Q.~V.; and Zhou, D. 2022.
\newblock Chain-of-{Thought} {Prompting} {Elicits} {Reasoning} in {Large}
  {Language} {Models}.
\newblock \emph{Advances in Neural Information Processing Systems}, 35:
  24824--24837.

\bibitem[{Wray, Kirk, and Laird(2021)}]{wray_language_2021}
Wray, R.~E.; Kirk, J.~R.; and Laird, J.~E. 2021.
\newblock Language {Models} as a {Knowledge} {Source} for {Cognitive} {Agents}.
\newblock In \emph{Proceedings of the {Ninth} {Annual} {Conference} on
  {Advances} in {Cognitive} {Systems}}. Virtual.

\bibitem[{Wu et~al.(2023)Wu, Antonova, Kan, Lepert, Zeng, Song, Bohg,
  Rusinkiewicz, and Funkhouser}]{wu_tidybot_2023}
Wu, J.; Antonova, R.; Kan, A.; Lepert, M.; Zeng, A.; Song, S.; Bohg, J.;
  Rusinkiewicz, S.; and Funkhouser, T. 2023.
\newblock {TidyBot}: {Personalized} {Robot} {Assistance} with {Large}
  {Language} {Models}.
\newblock Number: arXiv:2305.05658 arXiv:2305.05658 [cs].

\bibitem[{Yost(1993)}]{yost_acquiring_1993}
Yost, G.~R. 1993.
\newblock Acquiring {Knowledge} in {Soar}.
\newblock \emph{Intelligent Systems}, 8(3): 26--34.

\end{thebibliography}

\end{document}